\documentclass{article}

    \usepackage[main,final]{neurips_2025}

\usepackage[utf8]{inputenc} % allow utf-8 input
\usepackage[T1]{fontenc}    % use 8-bit T1 fonts
\usepackage{hyperref}       % hyperlinks
\usepackage{url}            % simple URL typesetting
\usepackage{booktabs}       % professional-quality tables
\usepackage{amsfonts}       % blackboard math symbols
\usepackage{nicefrac}       % compact symbols for 1/2, etc.
\usepackage{microtype}      % microtypography
\usepackage{xcolor}         % colors
\usepackage{amsmath, amssymb, amsthm}
\usepackage{tikz}
\usepackage{tcolorbox}
\tcbuselibrary{skins}
\usetikzlibrary{arrows.meta}
\usepackage{pgfplots}
\usepackage{enumitem}
\usepackage{wrapfig}
\pgfplotsset{compat=1.18}
\usepackage[utf8]{inputenc} % allow utf-8 input
\usepackage[capitalize]{cleveref}
\usepackage{bbding}
\usepackage{pifont}
\usepackage{colortbl}
\usepackage{soul}
\usepackage{multirow}
\usepackage[misc]{ifsym}
\crefname{section}{Sec.}{Secs.}

\definecolor{myyellow}{rgb}{0.96, 0.64, 0.38}
\definecolor{myblue}{rgb}{0, 0.7, 0.93}

\theoremstyle{plain}
\newtheorem{theorem}{Theorem}[section] % 定义定理环境，编号按节递增

\usepgfplotslibrary{groupplots}
\hypersetup{
    colorlinks=true,
    citecolor=green,
}
\title{Mitigating Overthinking in Large Reasoning Models via Manifold Steering}

\author{Yao Huang$^{1,2}$, Huanran Chen$^2$, Shouwei Ruan$^1$,\\\quad \textbf{Yichi Zhang$^2$, Xingxing Wei$^{1,4}$,  Yinpeng Dong$^{2,3\thanks{Corresponding Author}}$}\\$^1$Institute of Artificial Intelligence, Beihang University, Beijing 100191, China\\$^2$College of AI, Tsinghua University, Beijing 100084, China $^{3}$Shanghai Qi Zhi Institute\\
$^{4}$State Key Laboratory of Virtual Reality Technology and Systems, Beihang University\\\Letter\,: \small\texttt{\{y\_huang, xxwei\}@buaa.edu.cn, dongyinpeng@mail.tsinghua.edu.cn}}

\begin{document}

\maketitle

\begin{abstract}
Recent advances in Large Reasoning Models (LRMs) have demonstrated remarkable capabilities in solving complex tasks such as mathematics and coding. However, these models frequently exhibit a phenomenon known as \textit{overthinking} during inference, characterized by excessive validation loops and redundant deliberation, leading to substantial computational overheads. In this paper, we aim to mitigate overthinking by investigating the underlying mechanisms from the perspective of mechanistic interpretability. We first showcase that the tendency of overthinking can be effectively captured by a single direction in the model's activation space and the issue can be eased by intervening the activations along this direction. However, this efficacy soon reaches a plateau and even deteriorates as the intervention strength increases. We therefore systematically explore the activation space and find that the overthinking phenomenon is actually tied to a low-dimensional manifold, which indicates that the limited effect stems from the noises introduced by the high-dimensional steering direction. Based on this insight, we propose \textbf{Manifold Steering}, a novel approach that elegantly projects the steering direction onto the low-dimensional activation manifold given the theoretical approximation of the interference noise. Extensive experiments on DeepSeek-R1 distilled models validate that our method reduces output tokens by up to 71\% while maintaining and even improving the accuracy on several mathematical benchmarks. Our method also exhibits robust cross-domain transferability, delivering consistent token reduction performance in code generation and knowledge-based QA tasks. Code is available at: \href{https://github.com/Aries-iai/Manifold\_Steering}{https://github.com/Aries-iai/Manifold\_Steering}.
\end{abstract}
\section{Introduction}
Building on the versatility of Large Language Models (LLMs) in text generation, particularly their emergent ability in chain-of-thought (CoT) reasoning~\cite{wei2022chain}, the field is now undergoing a transition toward Large Reasoning Models (LRMs). Exemplified by the OpenAI o-series~\cite{openai2024learning} and the DeepSeek-R1 series~\cite{deepseekai2025deepseekr1incentivizingreasoningcapability}, LRMs acquire internal capabilities for long-horizon reasoning through reinforcement learning with verifiable rewards. These models are able to explore diverse solution paths, reflect on potential errors, refine intermediate steps, and validate final outputs, mimicking the process of human problem-solving by scaling inference-time computations~\cite{openai2024learning}. As a result, they excel in domains such as mathematics~\cite{ahn2024large, luo2023wizardmath, wang2025polymath} and coding~\cite{ni2024next, yang2025code, yu2024siam}. This makes them well-suited for tasks that demand deep logical analysis and paves the way for their applications in more complex scenarios, including web search~\cite{li2025webthinker} and research assistance~\cite{zheng2025deepresearcher}.

However, despite the remarkable reasoning capabilities, LRMs often suffer from a critical efficiency issue known as \emph{overthinking}~\cite{sui2025stop}, where they generate excessive and unnecessary reasoning steps, even for simple questions. For example, when tasked with a straightforward calculation, like ``$2 + 3$'' \cite{chen2024not}, an LRM might redundantly validate its approach or explore irrelevant alternatives, significantly increasing the computation overloads. This overthinking not only impacts inference latencies, posing great challenges for time-critical applications, but also risks degrading performance by entangling the model in repetitive verification loops or unproductive reasoning paths \cite{chen2024not, fu2024efficiently, wang2025thoughts}. To mitigate such overthinking in LRMs, several approaches~\cite{baek2025towards, chen2025seal, fu2024efficiently, liu2025thought} have recently been proposed. They often utilize external mechanisms to regulate reasoning and prevent overthinking, which can incur additional computational overhead for probing \cite{fu2024efficiently} or be susceptible to performance degradation due to the reliance on external models \cite{liu2025thought}. While these methods address overthinking from external and behavioral perspectives -- relying on human-designed workflows and interventions, the underlying mechanism remains underexplored, posing significant challenges to achieving intrinsic mitigation.

In this paper, we address the overthinking problem of LRMs through mechanistic interpretability~\cite{zou2023representation}, based on an in-depth analysis of their internal states. Specifically, we attribute this phenomenon to the distinctive activation patterns in the deeper layers of the model and identify a single, interpretable direction by comparing the differences in the activations between overthinking and concise reasoning. By manipulating the activations along this direction, we can effectively steer the model away from overthinking tendencies. However, this intervention is insufficient to fully resolve the problem. As in \cref{fig:cumulative_variance_ratio}(a), the reduction in output tokens does not consistently scale with increasing intervention strength. This suggests that the computed steering direction is not accurate enough and introduces unintended \textit{interference noise}.

To address this issue, we further analyze the model's activation patterns and find that the overthinking phenomenon is intrinsically tied to a low-dimensional manifold, which can be well approximated by a linear subspace. 
This result sheds light on why high-dimensional steering directions often introduce noises, as they fail to align with the underlying structure of model activations. To more effectively  mitigate overthinking, we introduce a \textbf{Manifold Steering} method to align the steering direction with the reconstructed low-dimensional manifold. We first theoretically derive a linear approximation of the amplitudes of the interference noise and then project the steering direction by nullifying this approximated term. In this way, we can effectively purify the steering direction and better mitigate the overthinking issue with larger intervention strength, as depicted in \cref{fig:cumulative_variance_ratio}(a).

Extensive experiments on multiple DeepSeek-R1 distilled models~\cite{deepseekai2025deepseekr1incentivizingreasoningcapability} of different sizes verify the effectiveness of our manifold steering method. We first test it on mathematical datasets of varying difficulty, including GSM8K~\cite{cobbe2021gsm8k}, Math500~\cite{lightman2023lets}, AMC2023~\cite{amc2023}, and AIME2024~\cite{aime2024}. Our method achieves up to 71\% tokens reduction while consistently maintaining or improving accuracy. Moreover, it exhibits robust cross-domain transferability, delivering consistent mitigation effects in tasks such as LiveCodeBench~\cite{jain2024livecodebench} (code generation) and Diamond-GPQA~\cite{rein2024gpqa} (knowledge-based QA), surpassing existing methods in both overthinking mitigation and accuracy preservation.

\section{Related Work}

\textbf{Mechanistic Interpretability.} Mechanistic interpretability~\cite{bereska2024mechanistic,conmy2023towards, nanda2023progress, panickssery2023steering, rai2024practical, subramani2022extracting, turner2023steering, zou2023representation} seeks to reverse-engineer the internal computations of LLMs to uncover the causal mechanisms underlying their behavior, offering fine-grained insights into learned representations and decision processes. A key technique within this framework involves identifying \textit{steering directions}~\cite{panickssery2023steering, subramani2022extracting, turner2023steering}—linear vectors in the activation space that correspond to specific model behaviors. By manipulating these directions during inference, researchers can precisely control outputs, such as ablating refusal behaviors in safety-critical scenarios~\cite{arditi2024refusal, yu2025robust}. Similarly, Cao et al.~\cite{cao2024personalized} proposed Bi-directional Preference Optimization (BiPO), leveraging steering vectors derived from contrasting human preference pairs to customize attributes like truthfulness and hallucination. These approaches highlight the versatility of steering directions in manipulating models' behaviors. Additionally, some efforts have explored dimensionality reduction in activation spaces: \cite{cancedda2024spectral} use SVD-based spectral filtering to suppress noise in residual streams, while others derive steering directions from probabilistic classification of user history for multi-preference alignment~\cite{song2025effectively}. Our work extends this paradigm to address \textit{overthinking} in LLMs~\cite{chen2024not, fu2024efficiently, sui2025stop}, a phenomenon characterized by redundant or divergent reasoning trajectories. By analyzing the latent space, we identify a steering direction that encapsulates overthinking and further propose \textit{manifold steering}, a novel method that projects this direction onto a low-dimensional manifold to mitigate interference noise, thereby improving its performance.

\textbf{Overthinking Mitigation.}
Efforts~\cite{ baek2025towards,  chen2025seal, fu2024efficiently, liu2025thought, pu2025thoughtterminator, yang2025dynamic} to mitigate overthinking in LRMs have gained traction as a means to enhance inference efficiency and output quality. 
Among them, the training-based method~\cite{pu2025thoughtterminator} tend to modify the reward function for length control in reinforcement learning. However, these incur significant computational costs and are orthogonal to inference-time interventions like ours, thus warranting no direct comparison in this work. Existing training-free methods mainly rely on external mechanisms to regulate reasoning. For instance, Dynasor~\cite{fu2024efficiently} employs periodic monitoring to detect and halt redundant reasoning, incurring computational overhead, while Thought Manipulation~\cite{liu2025thought} uses auxiliary models to guide inference, limited by the external model’s performance.  These shortcomings suggest that a more fundamental solution lies in understanding and modifying the model’s internal reasoning processes. 
Though some concurrent works~\cite{baek2025towards, chen2025seal} have tried to leverage mechanistic interpretability for achieving it, they only partially reduce overthinking, quickly encountering bottlenecks due to interference noise in high-dimensional steering directions. In contrast, we propose manifold steering to project the steering direction onto a low-dimensional manifold, effectively eliminates interference noise, achieving superior overthinking mitigation and substantial token reductions across diverse tasks, as demonstrated in \cref{sec:sota_compare}.

\section{Mechanistic Analysis of Overthinking}
\label{sec:mech}

In this section, we investigate the phenomenon of overthinking within the activation space of Large Reasoning Models (LRMs) and identify a general mechanism by which ablating a single direction in the activation space can reduce redundant reasoning steps to some extent.

\subsection{Background}
\label{sec:background}

\textbf{Transformers.}
Decoder-only transformer language models~\cite{liu2018generating, vaswani2017attention} map an input token sequence $\mathbf{x} = [x_1, \dots, x_T]$ to a probability distribution over the vocabulary for next-token prediction. Each token $x_i$ is associated with a sequence of residual stream activations $\mathbf{h}^{(l)}(x_i) \in \mathbb{R}^d$ across $L$ layers, initialized by the token embedding $\mathbf{h}^{(0)}(x_i) = \text{Embed}(x_i)$. At each layer $l \in \{1, \dots, L\}$, the residual stream $\mathbf{h}^{(l)}(x_i)$ is updated by combining the previous layer’s activation $\mathbf{h}^{(l-1)}(x_i)$ with two components: (\romannumeral 1) a multi-head self-attention mechanism, which computes $\mathbf{a}^{(l)}(x_{1:i})$ by attending to prior tokens $\{x_j : j \leq i\}$ using a causal mask to enforce autoregressive context flow; and (\romannumeral 2) a multi-layer perceptron (MLP), which applies non-linear transformations to the post-attention state $\mathbf{h}^{(l-1)}(x_i) + \mathbf{a}^{(l)}(x_{1:i})$ and produces $\mathbf{m}^{(l)}(x_i)$. The whole update is expressed as follows:
\begin{equation}
\mathbf{h}^{(l)}(x_i) = \mathbf{h}^{(l-1)}(x_i) + \mathbf{a}^{(l)}(x_{1:i}) + \mathbf{m}^{(l)}(x_i), \quad \mathbf{m}^{(l)}(x_i) = \text{MLP}(\mathbf{h}^{(l-1)}(x_i)+\mathbf{a}^{(l)}(x_{1:i})).
\label{act}
\end{equation}
Through autoregressive aggregation, each $\mathbf{h}^{(l)}(x_i)$ aggregates context from prior tokens, with the final token’s residual stream $\mathbf{h}^{(l)}(x) :\rightarrow \mathbf{h}^{(l)}(x_T)$ encapsulating the entire input’s context.

\textbf{Large Reasoning Models.}
LRMs are tailored for complex problem-solving and instruction-following, which leverage structured templates to handle user inputs:
\begin{center}
\begin{minipage}{0.9\textwidth}
\centering
\begin{tcolorbox}[colback=gray!7, width=\textwidth, boxrule=0.5pt, arc=2mm, boxsep=0.8pt, left=5pt, right=5pt]
\texttt{<|begin\_of\_sentence|><|User|>\{instruction\}<|Assistant|><think>\textbackslash n}
\end{tcolorbox}
\end{minipage}
\end{center}
where the content following \texttt{<think>\textbackslash n} comprises the model’s reasoning process and final answer, separated by \texttt{</think>}. Despite the excellent reasoning capabilities of these models, they often exhibit the overthinking phenomenon~\cite{chen2024not, cuadron2025danger} during the reasoning process, characterized by repetitive validation or redundant deliberation. As a high-level cognitive phenomenon, overthinking may manifest in the model’s residual stream activations, similar to other abstract concepts such as safety \cite{bereska2024mechanistic} and honesty \cite{zou2023representation}, as widely studied from the perspective of mechanistic interpretability. This suggests that overthinking and concise reasoning exhibit distinct activation patterns. In the next section, we systematically examine this hypothesis and investigate whether these activation differences are sufficient to identify a specific direction that characterizes overthinking -- one that, if isolated, could be ablated to improve reasoning efficiency.

\subsection{Extracting and Ablating an Overthinking Direction}
\label{sec:extraction}
Before extracting an overthinking direction, we first investigate whether the residual stream activations corresponding to redundant and concise reasoning are separable in the model’s activation space, as this is a necessary condition for identifying a meaningful and controllable direction. Drawing on \cite{fu2024efficiently}, we also focus on mathematical problems, where the overthinking phenomenon is particularly pronounced. To construct representative data, we first randomly sample questions from the OpenMathInstruct-2 training set~\cite{toshniwal2024openmathinstruct}. For each model, five responses per question are independently generated. Based on these responses, we construct two model-specific datasets\footnote{Details on data selection and dataset composition are provided in the \cref{sec:implement}.} as follows:
\begin{itemize}[leftmargin=1.5em]
    \item \textbf{Redundant set} $ D_{\text{redundant}} $: consists of questions for which all five responses exceed 16k tokens and contain hesitation keywords (e.g., ``\textit{wait}'', ``\textit{alternatively}'', etc.) surpassing a specified number.
    \item \textbf{Concise set} $ D_{\text{concise}} $: consists of questions for which all five responses are under 1k tokens and contain none of the hesitation keywords.
\end{itemize}

\begin{figure}[!t]
    \centering
    \includegraphics[width=\linewidth]{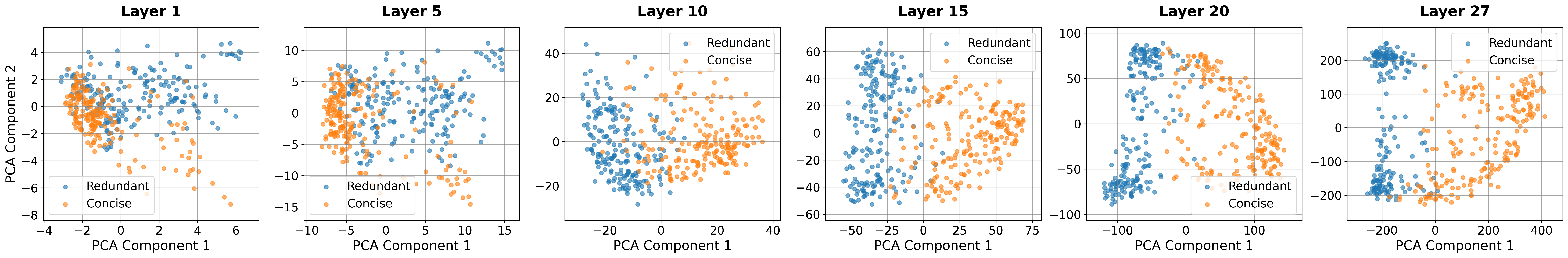}
    \caption{Visualization of residual stream activations $ \mathbf{h}^{(l)}(x) $ for $ D_{\text{redundant}} $ and $ D_{\text{concise}} $ across different layers of DeepSeek-R1-Distill-Qwen-7B (R1-7B). Early layers show considerable overlap between redundant and concise data, while middle-to-late layers exhibit distinct separation.}
    \label{fig:pca}
    % \vspace{-2ex}
\end{figure}

As demonstrated in \cref{fig:pca}, we visualize the distribution of residual stream activations $ \mathbf{h}^{(l)}(x) $ for both $ D_{\text{redundant}} $ and $ D_{\text{concise}} $ across different layers of R1-7B. We observe that, while early layers exhibit substantial overlap between the two distributions, the middle-to-late layers display clear separation. This separation indicates that the overthinking phenomenon is more prominent in specific layers and provides empirical support for identifying a meaningful overthinking direction.

We use the \textit{difference-in-means} technique~\cite{belrose2023diffinmeans} for extracting the steering direction, which computes the mean difference in residual stream activations between the redundant and concise data for each layer $l$. The overthinking direction $ \mathbf{r}^{(l)} $ is then defined as:
\begin{equation}
\mathbf{r}^{(l)} = \frac{1}{|D_{\text{redundant}}|} \sum_{x \in D_{\text{redundant}}} \mathbf{h}^{(l)}(x) - \frac{1}{|D_{\text{concise}}|} \sum_{x \in D_{\text{concise}}} \mathbf{h}^{(l)}(x),
\label{eq:overthinking_direction}
\end{equation}
where $ \mathbf{h}^{(l)}(x) $ denotes the residual stream activation of the final token of input $ x $ at layer $ l $, with $x$ being the prompt concatenated with the response for $x \in \mathcal{D}_{\text{redundant}}$ but only the prompt for $x \in \mathcal{D}_{\text{concise}}$. The direction $ \mathbf{r}^{(l)} $ is normalized to unit length, i.e., $ \mathbf{r}^{(l)} = \mathbf{r}^{(l)} / \|\mathbf{r}^{(l)}\|_2 $. Following \cite{arditi2024refusal}, we also select the single most effective direction $ \mathbf{r}^{(l^*)} $ and apply it for intervention across all layers.

Finally, to further explore the role of the overthinking direction $ \mathbf{r}^{(l^*)} $ in the model’s computations, we ablate the component aligned with $ \mathbf{r}^{(l^*)} $ each residual stream activation $ \mathbf{h} $. Specifically, the modified activation $ \mathbf{h}' $ is computed as:
\begin{equation}
\mathbf{h}' = \mathbf{h} - \alpha \times \mathbf{r}^{(l^*)} (\mathbf{r}^{(l^*)})^\top \mathbf{h},
\label{eq:ablation}
\end{equation}
where $ \alpha $ controls the intervention strength. We apply this ablation to every activation $ \mathbf{h}^{(l)}(x_i) $, across all layers $ l $ and token positions $ i $. The parameter $ \alpha $ allows adapting the extent of overthinking mitigation, balancing the reduction of redundant reasoning with the problem-solving accuracy.

\section{Manifold Steering for Robust Intervention}
Following our mechanistic analysis in \cref{sec:mech}, which identifies a single direction capturing overthinking in the model's activation space, we proceed to explore whether increasing the intervention strength $\alpha$ further reduces redundant reasoning, as expected.

\subsection{Low-Dimensional Manifold Analysis}
\label{sec:low_dime}

To rigorously evaluate the effect of increasing intervention strength $\alpha$ for the direction $\mathbf{r}^{(l^*)}$, derived via \cref{eq:overthinking_direction}, we test R1-7B's performance on the Math500 dataset~\cite{lightman2023lets}, a diverse mathematical test set, different from the anchor dataset used in \cref{sec:extraction}.

\begin{wrapfigure}{r}{4.5cm}
    \vspace{-3ex}
    \includegraphics[width=4.5cm]{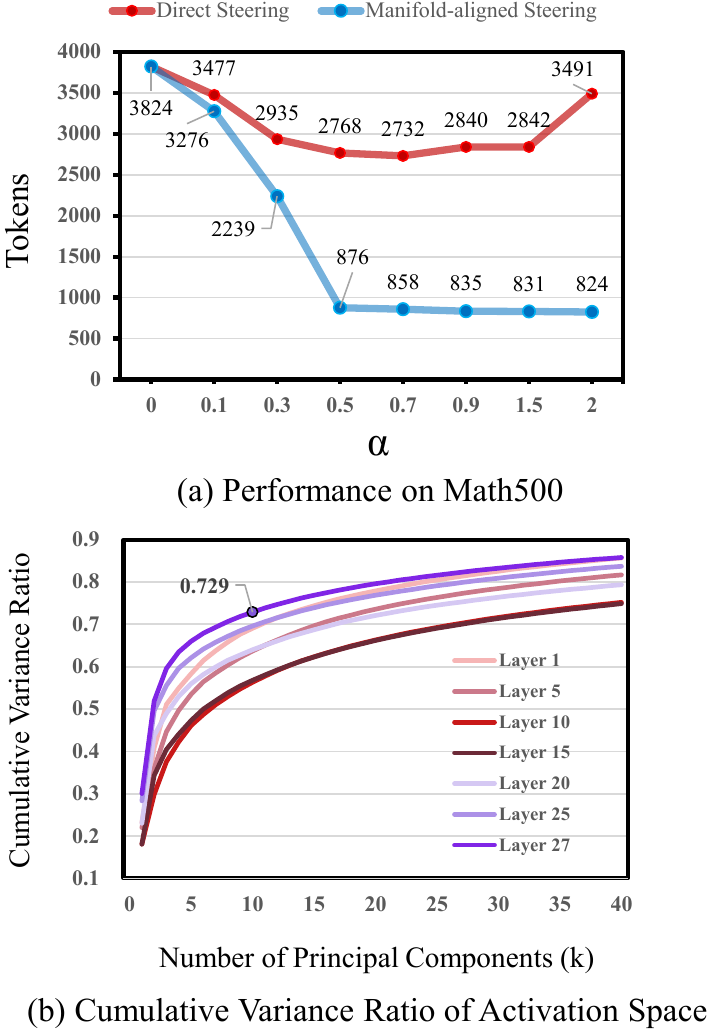}
    \caption{(a) Performance of R1-7B with varying $\alpha$ for direct and manifold steering on Math500. (b) Cumulative variance ratio of R1-7B's activation space on $D_{\text{redundant}}$ across different hidden layers.} 
    \label{fig:cumulative_variance_ratio}
    \vspace{-7ex}
\end{wrapfigure}

\textbf{Formulation of Interference Noise.} As illustrated in \cref{fig:cumulative_variance_ratio}(a), increasing the intervention strength $\alpha$ initially reduces the token count. However, beyond a certain threshold, the token count ceases to decrease, and as $\alpha$ continues to increase beyond $1.5$, it rebounds, even nearly returning to levels observed without intervention. This suggests that the intervention direction $\mathbf{r}^{(l^*)}$ may be imprecise and introduces unintended noise in the model's activation space, which is defined as \textit{interference noise}. Meanwhile, we confirm the model collapse caused by interference noise, as below:
\begin{center}
\begin{minipage}{\textwidth}
\centering
\begin{tcolorbox}[colback=gray!7, width=\textwidth, boxrule=0.5pt, arc=2mm, boxsep=0.8pt, left=5pt, right=5pt]
\texttt{\textbf{Instruction: }What power of 4 is equal to 8? Express your answer.}

\texttt{\textbf{Response: }10000000000000000000000 $\cdots$}
\end{tcolorbox}
\end{minipage}
\end{center}
Thus, we can hypothesize that the $\mathbf{r}^{(l^*)}$, computed via the difference-in-means method in $\mathbb{R}^d$, comprises both the overthinking direction $\mathbf{r}_{\textit{overthinking}}$ and an orthogonal\footnote{The orthogonality of $\mathbf{r}_{\textit{overthinking}}$ and $\mathbf{r}_{\textit{other}}$ is a property arising from the principles of PCA.} interference component $\mathbf{r}_{\textit{other}}$, such that $\mathbf{r}^{(l^*)} = \mathbf{r}_{\textit{overthinking}} + \mathbf{r}_{\textit{other}}$. The \cref{eq:ablation} actually modifies the activation as follows:
\begin{equation}
\mathbf{h}'^{(l)}(x_i) = \mathbf{h}^{(l)}(x_i) - \alpha \left[ \underbrace{(\mathbf{r}_{\textit{overthinking}})^\top \mathbf{h}^{(l)}(x_i) \mathbf{r}_{\textit{overthinking}}}_{\text{overthinking component}} + \underbrace{(\mathbf{r}_{\textit{other}})^\top \mathbf{h}^{(l)}(x_i) \mathbf{r}_{\textit{other}}}_{\text{interference component}} \right].
\label{eq:intervention_decomp}
\end{equation}
The $\mathbf{r}_{\textit{other}}$ term perturbs $\mathbf{h}'^{(l)}(x_i)$, potentially disrupting unrelated capabilities such as normal expression, especially for large $\alpha$. This interference explains the token count rebound beyond $\alpha=1.5$, as the intervention affects dimensions irrelevant to overthinking.

\textbf{Linear Low-Dimensional Manifold Verification.} As shown above, the direct computation of difference in high-dimensional activation space leads to noisy estimation due to the existence of the interference part $\mathbf{r}_\textit{other}$. A straightforward solution is to estimate the amplitude of $\mathbf{r}_\textit{other}$ and remove its influence from the steering direction $\mathbf{r}^{(l^*)}$. However, it is orthogonal to the overthinking direction $\mathbf{r}_{\textit{overthinking}}$ and is decided by the space of $\mathbf{r}_{\textit{overthinking}}$. Inspired by prior work~\cite{engels2024not} that the activations in LLMs reside on a low-dimensional manifold $\mathcal{M} \subset \mathbb{R}^d$, it is reasonable to assume that $\mathbf{r}_{\textit{overthinking}}$, representing the shift between activations of redundant and concise reasoning paths, also falls into this manifold. To verify this, we employ a simple linear method -- Principal Component Analysis (PCA), on the activations from the complete reasoning dataset $D_{\text{reasoning}} = D_{\text{redundant}}\bigcup D_{\text{concise}}$ at layer $l$. Let $\mathbf{A}^{(l)} = [\mathbf{h}^{(l)}(x_1), \dots, \mathbf{h}^{(l)}(x_N)] \in \mathbb{R}^{d \times N}$ denote the matrix of activation vectors $\mathbf{h}^{(l)}(x_i) \in \mathbb{R}^d$ for inputs $x_i \in D_{\text{reasoning}}$. We compute the covariance matrix and its eigendecomposition as:
\begin{equation}
    \mathbf{C}^{(l)} = \frac{1}{N-1} (\mathbf{A}^{(l)}-\bar{\mathbf{A}}^{(l)}) (\mathbf{A}^{(l)} - \bar{\mathbf{A}}^{(l)})^\top = \mathbf{U}^{(l)} \mathbf{\Lambda}^{(l)} (\mathbf{U}^{(l)})^\top, 
\label{eq:pca_formulas}
\end{equation}
where $\bar{\mathbf{A}}^{(l)} = \frac{1}{N} \sum_{i=1}^N \mathbf{h}^{(l)}(x_i)$, $\mathbf{U}^{(l)} \in \mathbb{R}^{d \times d}$ contains the principal components, $\mathbf{\Lambda}^{(l)} = \text{diag}(\lambda_1^{(l)}, \dots, \lambda_d^{(l)})$, and $\text{VR}(k) = \frac{\sum_{i=1}^k \lambda_i^{(l)}}{\sum_{i=1}^d \lambda_i^{(l)}}$ is the variance ratio. As \cref{fig:cumulative_variance_ratio}(b) shows, the top $k = 10$ components account for over 70\% of the variance, indicating that the effective dimension of $\mathcal{M}$, denoted $d_{\text{eff}}$, is significantly smaller than the ambient dimension $d$. This confirms the low-dimensional structure of $\mathcal{M}$. This also suggests that the linear manifold composed by the orthogonal basis effectively captures the activations of reasoning trajectories and therefore overthinking direction $\mathbf{r}_{\textit{overthinking}}$ can be estimated using simple linear dimensionality reduction in this subspace. Eventually, this finding supports our  earlier hypothesis (\cref{eq:overthinking_direction}) that the steering direction $\mathbf{r}^{(l^*)} = \mathbf{r}_{\textit{overthinking}} + \mathbf{r}_{\textit{other}}$ includes an orthogonal interference component $\mathbf{r}_{\textit{other}}$, which falls into $\mathcal{M}^\perp$.

\subsection{Theoretical Analysis of Interference Noise}
As discussed in \cref{sec:low_dime}, the overthinking phenomenon is tied to the low-dimensional manifold structure of the activation space. The overthinking direction $\mathbf{r}^{(l^*)}$, computed via \cref{eq:overthinking_direction}, introduces an orthogonal interference component $\mathbf{r}_{\textit{other}}$ due to computation in high-dimensional spaces. When the activation dimension $d$ far exceeds the sample size $N$ ($d \gg N$), interference noise even accumulates in $\mathcal{M}^\perp$, inflating $\mathbf{r}_{\textit{other}}$ and disrupting the model's other normal abilities. To further clarify the potential effects, we quantify this interference noise in the following theorem.
\begin{theorem}
(Proof in \cref{sec:prove}) Let $\mathbf{P}_{\mathcal{M}} = \mathbf{U}^{(l)}[:,1:k] (\mathbf{U}^{(l)}[:,1:k])^\top$ be the projection matrix onto the low-dimensional manifold $\mathcal{M}$, where $\mathbf{U}^{(l)}[:,1:k]$ contains top-$k$ principal components of the activation covariance $\mathbf{C}^{(l)}$ for $D_{\text{redundant}}$ and $D_{\text{concise}}$. As the sample sizes grow sufficiently large, the expected noise norm of $\mathbf{r}_{\textit{other}}$ is:
\begin{equation}
\mathbb{E}[\|\mathbf{r}_{\textit{other}}\|_2^2] = \text{tr}\left( (\mathbf{I} - \mathbf{P}_{\mathcal{M}}) \mathbf{\Sigma}_{\text{noise}}^{(l)} \right), \quad \mathbf{\Sigma}_{\text{noise}}^{(l)} = \frac{\mathbf{C}^{(l)}}{|D_{\text{redundant}}|} + \frac{\mathbf{C}^{(l)}}{|D_{\text{concise}}|}.
\label{eq:noise_norm}
\end{equation}
The trace is significant, indicating that the interference noise is substantial and is greatly likely to disrupt the model's normal abilities.
\label{thm:error_amplification}
\end{theorem}

The significant noise norm of $\mathbf{r}_{\textit{other}}$, as established in \cref{thm:error_amplification}, suggests that interventions using $\mathbf{r}^{(l^*)}$ introduce considerable perturbations in $\mathcal{M}^\perp$. Moreover, these perturbations can propagate through layers, amplified by attention mechanisms, non-linear activations, and residual connections, leading to more substantial shifts in the activation distribution. To understand this, we further analyze the mean activation shift and its layer-wise amplification in the following theorem.
\begin{theorem}
(Proof in \cref{sec:prove}) Let $\mathbf{r}^{(l^*)}$ and $\mathbf{r}_{\textit{other}}$ be as in \cref{thm:error_amplification}, and let the intervention be applied as in \cref{eq:intervention_decomp}. The mean activation shift at layer $l$ and its amplification through layers are:
\begin{align}
\Delta \boldsymbol{\mu}^{(l)} &= -\alpha \frac{1}{N} \sum_{i=1}^N [(\mathbf{r}^{(l^*)})^\top \mathbf{h}^{(l)}(x_i)] \mathbf{r}^{(l^*)}, \quad \|\Delta \boldsymbol{\mu}^{(l)}\|_2 \propto \alpha \|\mathbf{r}_{\textit{other}}\|_2, \label{eq:activation_shift} \\
\|\Delta \boldsymbol{\mu}^{(l+1)}\|_2 &\geq \gamma \|\Delta \boldsymbol{\mu}^{(l)}\|_2 + \alpha \gamma_{\text{attn}} \gamma_{\sigma} \sigma_{\text{min}}(\mathbf{W}^{(l+1)}) |(\mathbf{r}_{\textit{other}})^\top \mathbf{h}^{(l)}(x_i)| \|\mathbf{r}_{\textit{other}}\|_2, \label{eq:layer_amplification}
\end{align}
where $\alpha$ is the intervention strength, $\mathbf{h}^{(l)}(x_i)$ is the activation at layer $l$, $\mathbf{W}^{(l+1)}$ denotes the combined MLP and attention weights, $\gamma_{\text{attn}}$ and $\gamma_{\sigma}$ are the minimum amplification factors of the attention softmax and GeLU non-linearities, $\sigma_{\text{min}}(\mathbf{W}^{(l+1)})$ is the minimum singular value of the weights, and $\gamma > 1$ is the layer-wise amplification factor. 
\label{thm:activation_shift}
\end{theorem}
The shift in $\mathcal{M}^\perp$ is significant and grows through layer-wise propagation, driven by attention and non-linear transformations, severely disrupting the model's other normal abilities.
\subsection{Manifold Steering}

The interference direction $\mathbf{r}_{\textit{other}}$, quantified in \cref{thm:error_amplification}, causes activation shifts that amplify through transformer layers and disrupt reasoning (\cref{thm:activation_shift}). To eliminate this interference, a simple but effective approach is to set \cref{eq:noise_norm} to $\mathbf{0}$.
Based on this insight, we propose \textbf{Manifold Steering}, which projects the direction $\mathbf{r}^{(l^*)}$ onto $\mathcal{M}$ to mitigate $\mathbf{r}_{\textit{other}}$.

Formally, let $\mathbf{U}_{\text{eff}}^{(l)} \in \mathbb{R}^{d \times k}$ denote the top-$k$ principal components of the activation covariance in \cref{eq:pca_formulas}, spanning $\mathcal{M}$. The manifold direction is obtained by:
\begin{equation}
\mathbf{r}_{\textit{overthinking}}^{(l^*)} = \mathbf{P}_{\mathcal{M}} \mathbf{r}^{(l^*)} = \mathbf{U}_{\text{eff}}^{(l)} (\mathbf{U}_{\text{eff}}^{(l)})^\top \mathbf{r}^{(l^*)}, \quad \mathbf{r}_{\textit{overthinking}}^{(l)} = \frac{\mathbf{r}_{\textit{overthinking}}^{(l)}}{\|\mathbf{r}_{\textit{overthinking}}^{(l)}\|_2},
\label{eq:manifold_projection}
\end{equation}
where $\mathbf{P}_{\mathcal{M}} = \mathbf{U}_{\text{eff}}^{(l)} (\mathbf{U}_{\text{eff}}^{(l)})^\top$ is the projection matrix onto $\mathcal{M}$. The reason why the interference norm $\mathbb{E}[\|\mathbf{r}_{\textit{other}}\|_2^2] = \text{tr}\left( (\mathbf{I} - \mathbf{P}_{\mathcal{M}}) \boldsymbol{\Sigma}_{\text{noise}}^{(l)} \right) = 0$ holds is because $\boldsymbol{\Sigma}_{\text{noise}}^{(l)}$ is now primarily supported in $\mathcal{M}$, resulting in zero components under $I-\mathbf{P}_{\mathcal{M}}$, \textit{i.e.}, $\mathcal{M}^\perp$, successfully eliminating perturbations.
\begin{equation}
\mathbf{h}'^{(l)}(x) = \mathbf{h}^{(l)}(x) - \alpha \times\mathbf{r}_{\textit{overthinking}}^{(l)}(\mathbf{r}_{\textit{overthinking}}^{(l)})^\top \mathbf{h}^{(l)}(x_i).
\label{eq:aligned_intervention}
\end{equation}
The performance of manifold steering is shown in \cref{fig:cumulative_variance_ratio}(b) and \cref{sec:sota_compare}, where we find that, unlike the original paradigm, our manifold steering enables a sustained reduction in token count.

\section{Experiments}

\subsection{Experimental Setups}
\label{sec:setting}
We begin by briefly outlining the baseline methods, target LRMs, evaluation datasets, and metrics. For more
detailed descriptions of the experimental settings, please refer to \cref{sec:implement}.

\textbf{Baseline Methods.} 
We compare our manifold steering with two latest baselines, including Dynasor~\cite{fu2024efficiently} and SEAL~\cite{chen2025seal}, both chosen for their ability to maintain the model’s original accuracy. For their settings, we both adopt its official setting. To be aware, Dynasor's early stopping often omits the problem-solving process in the final answer, which is impractical for real-world applications. Thus, we require the model to provide a complete solution upon stopping.

\begin{table}[!t]
  \caption{Performance of Manifold Steering compared to Vanilla, Dynasor, and SEAL on GSM8K, MATH500, AMC2023, and AIME2024 for varied LRMs. Metrics include \textbf{Pass@1} ($\uparrow$) and \textbf{\#Tokens} ($\downarrow$). Changes relative to Vanilla are shown in \textcolor{myyellow}{yellow} for Pass@1 and \textcolor{myblue}{blue} for \#Tokens.}
  \vspace{-1ex}
  \label{tab:main}
  \resizebox{\linewidth}{!}{
  \begin{tabular}{c|c|cc|cc|cc|cc}
  \toprule[1.5pt]
                            &                                     & \multicolumn{2}{c}{\textbf{GSM8k}}                      & \multicolumn{2}{c}{\textbf{MATH500}} & \multicolumn{2}{c}{\textbf{AMC2023}} & \multicolumn{2}{c}{\textbf{AIME2024}} \\
  \multirow{-2}{*}{\textbf{Model}}   & \multirow{-2}{*}{\textbf{Methods}}           &  \textbf{Pass@1 ($\uparrow$, \%)}  & \multicolumn{1}{c}{\textbf{\#Tokens ($\downarrow$)}} &    \textbf{Pass@1 ($\uparrow$, \%)}      & \multicolumn{1}{c}{\textbf{\#Tokens ($\downarrow$)}}         & \textbf{Pass@1 ($\uparrow$, \%)}      & \multicolumn{1}{c}{\textbf{\#Tokens ($\downarrow$)}}      & \textbf{Pass@1 ($\uparrow$, \%)}       & \multicolumn{1}{c}{\textbf{\#Tokens ($\downarrow$)}}      \\\midrule[1.5pt]
  & Vanilla  &   76.7        &    2035      &     76.4      & 4762      &        70.0     &       7089        &      26.7        &    11352           \\
                            & Dynasor        &   77.1 (\textcolor{myyellow}{+0.4})        &   1035  (\textcolor{myblue}{-49\%})       & 77.2  (\textcolor{myyellow}{+0.8})          &    3694  (\textcolor{myblue}{-22\%})           &   72.5  (\textcolor{myyellow}{+2.5})      &       6505  (\textcolor{myblue}{-8\%})          &       26.7  (\textcolor{myyellow}{+0.0})        &     10564  (\textcolor{myblue}{-7\%})            \\
                      & SEAL        &  76.9  (\textcolor{myyellow}{+0.2})        &   1076  (\textcolor{myblue}{-47\%})        & 77.8  (\textcolor{myyellow}{+1.4})          &       3721  (\textcolor{myblue}{-22\%})           &      70.0  (\textcolor{myyellow}{+0.0})      &       6418  (\textcolor{myblue}{-10\%})          &       26.7  (\textcolor{myyellow}{+0.0})        &    10437  (\textcolor{myblue}{-8\%})        \\
         \multirow{-4}{*}{\textbf{R1-1.5B}} & \cellcolor{gray!10} Ours    &  \cellcolor{gray!10}\textbf{77.2  (\textcolor{myyellow}{+0.5})}        &   \cellcolor{gray!10}\textbf{593  (\textcolor{myblue}{-71\%})}        & \cellcolor{gray!10}\textbf{78.6  (\textcolor{myyellow}{+2.2})}  &       \cellcolor{gray!10}\textbf{3458  (\textcolor{myblue}{-27\%})}           &      \cellcolor{gray!10}\textbf{72.5  (\textcolor{myyellow}{+2.5})}      &     \cellcolor{gray!10}\textbf{6236  (\textcolor{myblue}{-12\%})}          &       \cellcolor{gray!10}\textbf{30.0  (\textcolor{myyellow}{+3.3})}        &  \cellcolor{gray!10}\textbf{10134  (\textcolor{myblue}{-11\%})}           \\\midrule\midrule
  & Vanilla  &     87.5          &   1143       &     88.2     &  3824                &      87.5       &        5871       &     50.0         &    10784            \\
                            & Dynasor              &   87.6  (\textcolor{myyellow}{+0.1})        & 732  (\textcolor{myblue}{-36\%})        & 88.2  (\textcolor{myyellow}{+0.0})          &        2723  (\textcolor{myblue}{-29\%})           &     85.0  (\textcolor{myyellow}{-2.5})      &      5121  (\textcolor{myblue}{-13\%})          &       46.7  (\textcolor{myyellow}{-3.3})        &    9864  (\textcolor{myblue}{-9\%})      \\
                            &SEAL    &   87.7  (\textcolor{myyellow}{+0.2})        &  829  (\textcolor{myblue}{-32\%})        & 87.8  (\textcolor{myyellow}{-0.4})          &        2651  (\textcolor{myblue}{-34\%})           &      85.0  (\textcolor{myyellow}{-0.0})      &       4750  (\textcolor{myblue}{-19\%})          &       46.7  (\textcolor{myyellow}{-3.3})       &     9394  (\textcolor{myblue}{-13\%})          \\
      \multirow{-4}{*}{\textbf{R1-7B}} & \cellcolor{gray!10} Ours    &   \cellcolor{gray!10}\textbf{87.6  (\textcolor{myyellow}{+0.1})}        &   \cellcolor{gray!10}\textbf{440  (\textcolor{myblue}{-62\%})}        & \cellcolor{gray!10}\textbf{88.4  (\textcolor{myyellow}{+0.2})}          &       \cellcolor{gray!10}\textbf{2239  (\textcolor{myblue}{-42\%})}           &      \cellcolor{gray!10}\textbf{87.5  (\textcolor{myyellow}{+0.0})}      &     \cellcolor{gray!10}\textbf{4440  (\textcolor{myblue}{-24\%})}          &       \cellcolor{gray!10}\textbf{53.3  (\textcolor{myyellow}{+3.3})}        &    \cellcolor{gray!10}\textbf{8457  (\textcolor{myblue}{-22\%})}           \\\midrule\midrule
  & Vanilla  &      82.7       &    1217      &     87.8     &    4009              &        85.0     &         5723      &      33.3        &    11278          \\
                        & Dynasor        &   82.9  (\textcolor{myyellow}{+0.2})        &   826  (\textcolor{myblue}{-32\%})        &  88.0  (\textcolor{myyellow}{+0.2})          &    3171  (\textcolor{myblue}{-21\%})           &     82.5  (\textcolor{myyellow}{-2.5})      &       5019  (\textcolor{myblue}{-12\%})          &       46.7  (\textcolor{myyellow}{+13.4})        &     9901  (\textcolor{myblue}{-12\%})            \\
                      & SEAL        &    82.7 (\textcolor{myyellow}{+0.0})        &    749 (\textcolor{myblue}{-38\%})        & 87.4 (\textcolor{myyellow}{-0.4})          &        3091 (\textcolor{myblue}{-23\%})           &      85.0 (\textcolor{myyellow}{+0.0})      &       4731 (\textcolor{myblue}{-17\%})          &       46.7 (\textcolor{myyellow}{+13.4})        &     9789 (\textcolor{myblue}{-13\%})        \\
        \multirow{-4}{*}{\textbf{R1-8B}} & \cellcolor{gray!10} Ours    &   \cellcolor{gray!10}\textbf{82.8  (\textcolor{myyellow}{+0.1})}        &   \cellcolor{gray!10}\textbf{542  (\textcolor{myblue}{-55\%})}        & \cellcolor{gray!10}\textbf{88.0  (\textcolor{myyellow}{+0.2})}          &       \cellcolor{gray!10}\textbf{2873  (\textcolor{myblue}{-29\%})}           &      \cellcolor{gray!10}\textbf{85.0  (\textcolor{myyellow}{+0.0})}      &     \cellcolor{gray!10}\textbf{4400  (\textcolor{myblue}{-23\%})}          &       \cellcolor{gray!10}\textbf{50.0  (\textcolor{myyellow}{+16.7})}        &    \cellcolor{gray!10}\textbf{9457  (\textcolor{myblue}{-16\%})}           \\\midrule\midrule
  & Vanilla  &     93.2       &   742        &      92.8    &       3496              &      90.0       &       5484        &      66.7        &            9986   \\
                     & Dynasor        &   93.4 (\textcolor{myyellow}{+0.2})        &  596  (\textcolor{myblue}{-20\%})        & 92.6 (\textcolor{myyellow}{-0.2})          &        3233  (\textcolor{myblue}{-8\%})           &     92.5 (\textcolor{myyellow}{+2.5})      &      4817  (\textcolor{myblue}{-12\%})          &      63.3 (\textcolor{myyellow}{-3.4})        &   8941  (\textcolor{myblue}{-11\%})            \\
                      & SEAL        &    93.6 (\textcolor{myyellow}{+0.2})        &    583 (\textcolor{myblue}{-21\%})        & 92.8 (\textcolor{myyellow}{+})          &        3139 (\textcolor{myblue}{-10\%})           &      87.5 (\textcolor{myyellow}{-2.5})      &       4470 (\textcolor{myblue}{-18\%})          &       60.0 (\textcolor{myyellow}{-6.7})        &     8563 (\textcolor{myblue}{-14\%})        \\
     \multirow{-4}{*}{\textbf{R1-14B}} & \cellcolor{gray!10} Ours    &   \cellcolor{gray!10}\textbf{93.6  (\textcolor{myyellow}{+0.4})}        &   \cellcolor{gray!10}\textbf{438  (\textcolor{myblue}{-41\%})}        & \cellcolor{gray!10}\textbf{92.8  (\textcolor{myyellow}{+0.0})}          &       \cellcolor{gray!10}\textbf{2074  (\textcolor{myblue}{-41\%})}           &     \cellcolor{gray!10}\textbf{90.0  (\textcolor{myyellow}{+0.0})}      &     \cellcolor{gray!10}\textbf{4061  (\textcolor{myblue}{-26\%})}          &       \cellcolor{gray!10}\textbf{63.3  (\textcolor{myyellow}{-3.4})}        &     \cellcolor{gray!10}\textbf{8132  (\textcolor{myblue}{-19\%})}           \\\bottomrule[1.5pt]
  \end{tabular}}
  \vspace{-0ex}
\end{table}

\begin{figure}[!t]
    \centering
    \includegraphics[width=\linewidth]{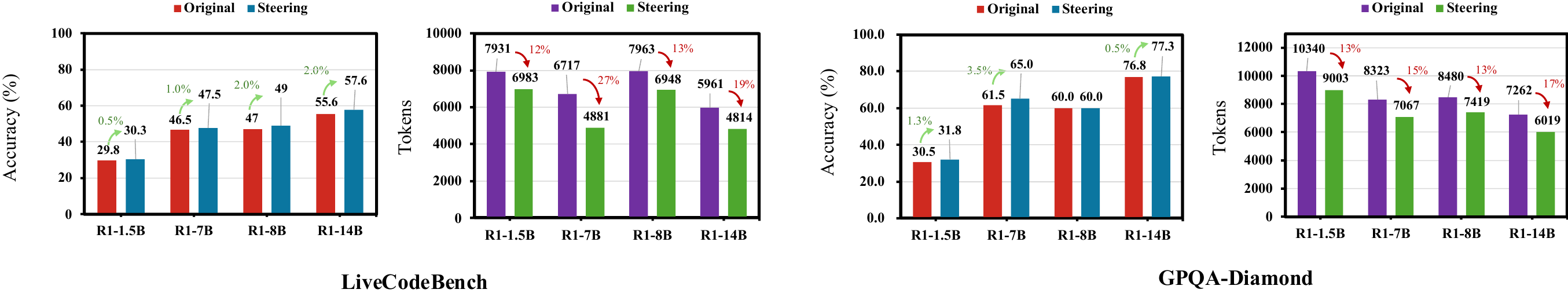}
    % \vspace{-3ex}
    \caption{Cross-domain performance of Manifold Steering for overthinking mitigation on LiveCodeBench (code generation) and GPQA-Diamond (disciplinary knowledge).}
\label{fig:cross_domain}
    \label{fig:corss_domain}
    % \vspace{-2ex}
\end{figure}

\textbf{Target LRMs.} For a comprehensive evaluation, we select the DeepSeek-R1-Distilled series~\cite{deepseekai2025deepseekr1incentivizingreasoningcapability}, comprising models of varying scales and architectures: DeepSeek-R1-Distill-Qwen (R1-1.5B, R1-7B, R1-14B) and DeepSeek-R1-Distill-Llama-8B (R1-8B). All models use recommended settings: temperature of 0.6, top-p of 0.95, and a maximum token limit of 16,384. 

\textbf{Evaluation Datasets \& Metrics.} To evaluate the effectiveness of Manifold Steering, we include mathematical datasets of varying difficulty: GSM8K~\cite{cobbe2021gsm8k}, MATH500~\cite{lightman2023lets}, AMC2023~\cite{amc2023}, and AIME2024~\cite{aime2024}. To further verify the transferability, we use LiveCodeBench~\cite{jain2024livecodebench} for code generation and GPQA-Diamond~\cite{rein2024gpqa} for expert-level disciplinary knowledge. All datasets are evaluated using \textbf{Pass@1} as the task-solving metric and the average token count (\textbf{\#Tokens}) for overthinking mitigation.

\textbf{Implementation Details.} The data for computing steering directions is filtered using the method outlined in \cref{sec:mech} on the OpenMathInstruct2 dataset~\cite{toshniwal2024openmathinstruct}. For each model, we specify the layer used to compute the steering direction and the intervention strength $\alpha$ as follows: R1-1.5B (layer 27, $\alpha = 0.7$), R1-7B (layer 27, $\alpha = 0.3$), R1-8B (layer 31, $\alpha = 0.5$), and R1-14B (layer 47, $\alpha = 0.3$). During inference, this direction is applied to all layers as stated in \cref{eq:ablation}.

\subsection{Performance of Manifold Steering in Overthinking Mitigation}
\label{sec:sota_compare}

We conduct experiments on four mathematical datasets of varying difficulty using four LRMs with different parameter sizes and architectures. \cref{tab:main} presents the results, where models are evaluated for accuracy and redundancy reduction. Based on  \cref{tab:main}, we draw the following observations:

\textbf{Manifold steering achieves the best performance across all models and datasets.}  Our method consistently outperforms baselines on four out-of-distribution mathematical datasets, with particularly strong results on GSM8K, where it achieves token reduction of 41\% $\sim$ 71\% while preserving accuracy. This is reasonable, as unlike Dynasor, which relies on external monitoring of the model's certainty, our method modifies model outputs at the more fundamental feature level to mitigate overthinking behavior. Moreover, Dynasor's reliance on external monitoring brings extra computational overhead. We include a comparison of average time cost to demonstrate this drawback in \cref{sec:compute}, while our method incurs nearly no latency. Compared to SEAL, our manifold steering effectively reduces interference noise, yielding a more precise direction. Additionally, we observe that the overthinking phenomenon diminishes to some extent as model parameter size increases, which is expected, as some overthinking stems from models’ inability to solve complex problems.

\begin{figure}[!t]
    \centering
    \includegraphics[width=\linewidth]{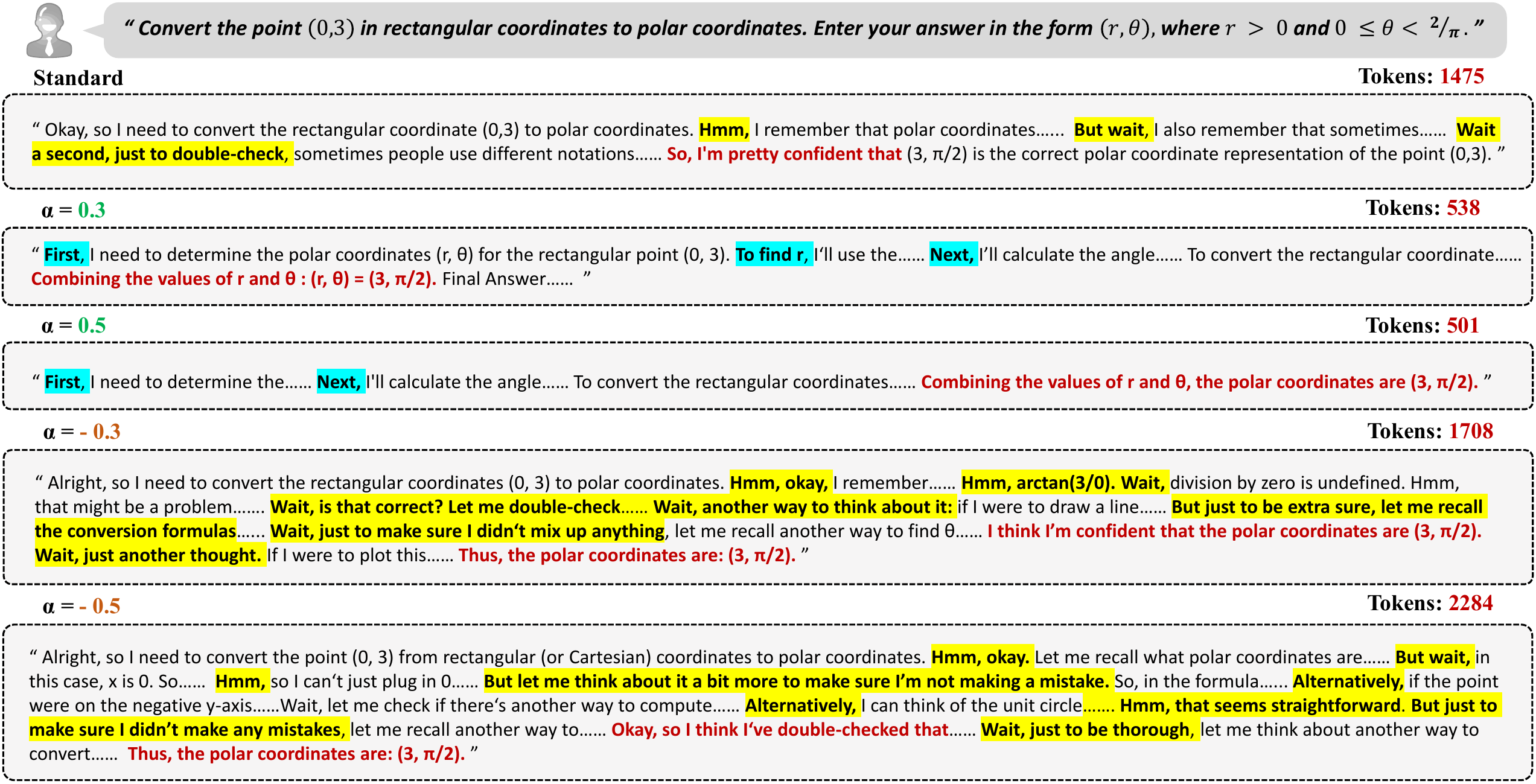}
    \vspace{-3ex}
    \caption{ An example of steering overthinking in model outputs. Forward steering yields concise, confident responses, eliminating hesitant phrases, while reverse steering induces verbose outputs.}
    \label{fig:case_study}
\end{figure}

\textbf{Overthinking is more pronounced in simpler problems.} 
As presented in \cref{tab:main}, all methods exhibit more effective overthinking mitigation on simpler datasets, with GSM8K and MATH500 ($\sim$ 40\%) showing greater token reduction compared to the more complex AMC2023 and AIME2024 datasets ($\sim$ 20\%). This suggests that overthinking is more pronounced in simpler problems, which is reasonable, as complex problems inherently require larger token budgets and may exceed the models' internal capabilities, thereby constraining mitigation effectiveness.

\subsection{Cross-Domain Transferability for Overthinking Mitigation}
To further investigate the transferability of manifold steering for overthinking mitigation, we assess its performance across two distinct domains: code generation and discipline-specific knowledge, both separate from the mathematical domain used for steering direction extraction. We utilize two representative datasets: \textbf{1) LiveCodeBench}~\cite{jain2024livecodebench}, a benchmark of coding challenges that probe algorithmic and programming expertise, and \textbf{2) GPQA-Diamond}~\cite{rein2024gpqa}, a carefully curated dataset of challenging multiple-choice questions targeting expert-level disciplinary knowledge across various fields. As shown in \cref{fig:corss_domain}, our manifold steering achieves token reduction of 12\% $\sim$ 27\% across both datasets while maintaining accuracy, demonstrating the generalizability of manifold steering to diverse domains. This cross-domain effectiveness offers multiple benefits: it incurs no additional computational overhead, adapts seamlessly to varied problem structures, and effectively mitigates overthinking without requiring domain-specific fine-tuning.

\subsection{Directional Analysis and Hyperparameter Tuning}
In this section, we first explore the precise impact of the direction computed by manifold steering on model outputs through case studies, analyzing how steering and its reversal affect response characteristics to better understand directionality's role. Then, we conduct hyperparameter tuning.

\textbf{Directional Analysis.} As shown in \cref{fig:case_study}, applying the steering direction for overthinking mitigation leads to model outputs that are significantly more concise and confident. Specifically, overthinking behaviors, such as hesitant phrases (e.g., ``wait''), frequent shifts in reasoning (e.g., ``alternatively''), and repetitive self-checking, are largely eliminated. The model generates streamlined responses with clear, focused reasoning, delivering direct outputs. This effect strengthens as the intervention strength increases to $ 0.5$. In contrast, when reverse steering is applied, the model becomes markedly more hesitant, often repeatedly checking. This leads to verbose outputs filled with excessive caution. Thus, it is crucial to underscore the role of directionality for overthinking.

\begin{wrapfigure}{r}{4.5cm}
    \vspace{-2ex}
    \includegraphics[width=4.5cm]{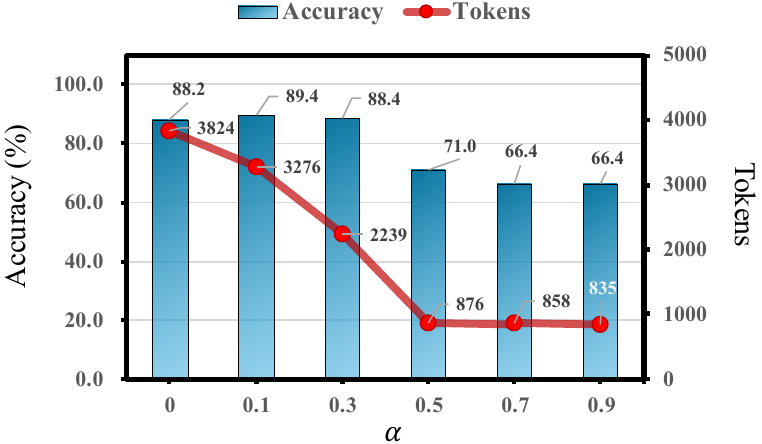}
    \caption{Hyperparameter tuning for strength $\alpha$ in R1-7B} 
    \label{fig:hyper}
    \vspace{-2ex}
\end{wrapfigure}
\textbf{Hyperparameter Tuning.} We use R1-7B model on MATH500 for this analysis, with results for other models in \cref{sec:hyper_more}. As shown in \cref{fig:hyper}, our manifold steering direction demonstrates efficacy at a much lower strength of $\alpha = 0.1$. As $\alpha$ increases, token counts continue to decrease, with a remarkable 77.1\% reduction observed at $\alpha = 0.5$. This substantial token reduction highlights the purity and effectiveness of our steering direction in mitigating overthinking. However, excessively rapid reasoning, induced by intervention strengths, can hinder the model's ability to thoroughly address complex problems, a phenomenon also observed in human cognition, leading to a decline in accuracy. To balance the trade-off between overthinking mitigation and maintaining accuracy, we select an intervention strength of $\alpha = 0.3$ as the optimal value for robust performance.

\begin{wrapfigure}{r}{4cm}
    \vspace{-4ex}
    \includegraphics[width=4cm]{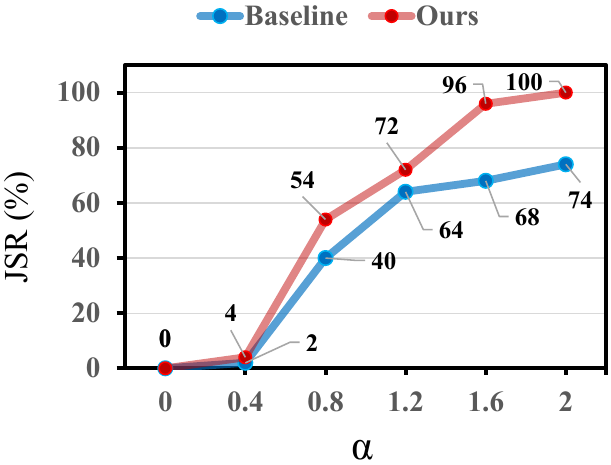}
    \caption{JSR of baseline and ours on Advbench.} 
    \label{fig:attack}
    \vspace{-2ex}
\end{wrapfigure}
\subsection{Cross-Task Transferability of Manifold Steering}
In this section, we investigate the applicability of manifold steering to tasks beyond overthinking mitigation, such as refusal feature ablation, to assess its cross-task transferability. Prior studies \cite{arditi2024refusal, yu2025robust} demonstrate that while steering directions can suppress refusal features in models, some instances persist unless intervention strength is increased, which risks model collapse. Here, we apply our manifold steering method using the Qwen2.5-7B-Instruct as the target LLM, computing the steering direction with the same data as in \cite{arditi2024refusal}. As shown in \cref{fig:attack}, our method achieves a 100\% jailbreak success rate (JSR) on AdvBench~\cite{zou2023universal} while the baseline~\cite{arditi2024refusal} obtains a JSR of 74\% ($\alpha = 2.0$), with all responses verified as valid through manual check, which further validates the robust transferability of manifold steering across diverse tasks and underscores the urgent need for enhanced safety efforts~\citep{dai2023safe, huang2025breaking, zhang2024multitrust, zhang2025realsafe, zhang2025stair} to ensure responsible AI.

\section{Discussion and Limitations}
\label{sec:discuss}
Our proposed manifold steering method has demonstrated robust effectiveness in mitigating overthinking, as evidenced by significant token reductions across varying LRMs. However, its applicability to multi-modal large language models remains unexplored. Additionally, while our approach excels in controlling overthinking with minimal accuracy trade-offs, its interaction with highly specialized tasks, such as domain-specific reasoning, e.g., legal or medical analysis, warrants further investigation. Moreover, the sensitivity of our method to varying intervention strengths suggests potential for optimizing dynamic steering strategies, where the strength adapts to task complexity in real-time. 
\section{Conclusion}
In this work, we propose manifold steering, a novel method to address overthinking in LRMs while preserving task performance without additional computational cost. Specifically, by aligning the steering direction with the low-dimensional activation manifold, our approach effectively eliminates the interference noise based on theoretical analysis. Extensive experiments across diverse models and datasets confirm substantial token reductions and robust cross-task transferability. These findings underscore the potential of manifold steering to enhance model efficiency and adaptability, opening new avenues for improving LRMs.

\ack
This work was supported by NSFC Projects (62576020, 62276149) and was also supported by the Fundamental Research Funds for the Central Universities.

\bibliographystyle{plain}
\bibliography{ref}

\newpage
\appendix

\section{Implement Details}
\label{sec:implement}
In this section, we will introduce the implementation details of our approach, focusing on the data selection process for computing the steering direction to mitigate overthinking in each model and the specific configurations of the baseline methods used for comparison.
\subsection{Data Selection for Direction Computation}
To construct representative datasets for computing the steering direction to mitigate overthinking, we begin by randomly sampling 20k questions from the OpenMathInstruct-2 training set~\cite{toshniwal2024openmathinstruct}. For each model, we generate five independent responses per question using the official sampling configuration: a temperature of 0.6, top-p of 0.95, and a maximum length of 16k tokens. These responses form the basis for constructing two model-specific datasets, the \textbf{Redundant set} ($D_{\text{redundant}}$) and the \textbf{Concise set} ($D_{\text{concise}}$), as described below:

\begin{itemize}[leftmargin=1.5em]
    \item \textbf{Redundant set ($D_{\text{redundant}}$):} This dataset includes questions where all five responses exceed 16k tokens without terminating and contain more than 20 times of hesitation keywords (e.g., ``wait'', ``alternatively'', etc.). To capture meaningful overthinking behavior, we process the responses using the following template, truncating the response at the occurrence of the hesitation keyword:

    \begin{center}
    \begin{minipage}{0.9\textwidth}
    \centering
    \begin{tcolorbox}[colback=gray!7, width=\textwidth, boxrule=0.5pt, arc=2mm, boxsep=0.8pt, left=5pt, right=5pt]
    \texttt{<|begin\_of\_sentence|><|User|>\{instruction\}<|Assistant|><think>\textbackslash n
    \{partial\_response\}\{hesitation keyword\}}
    \end{tcolorbox}
    \end{minipage}
    \end{center}

     The truncation at a hesitation keyword is reasonable because overthinking typically emerges after a certain point in the response, rather than immediately upon encountering the question. Moreover, through activation visualization, we observe no significant differences in the activation patterns of different hesitation keywords. Thus we choose ``wait'' as a consistent marker here.

    \item \textbf{Concise set ($D_{\text{concise}}$):} This dataset includes questions where all five responses are under 1k tokens and contain none of the hesitation keywords. The template for these responses includes only the instruction without the response, as they inherently represent concise and focused outputs:

    \begin{center}
    \begin{minipage}{0.9\textwidth}
    \centering
    \begin{tcolorbox}[colback=gray!7, width=\textwidth, boxrule=0.5pt, arc=2mm, boxsep=0.8pt, left=5pt, right=5pt]
    \texttt{<|begin\_of\_sentence|><|User|>\{instruction\}<|Assistant|><think>\textbackslash n}
    \end{tcolorbox}
    \end{minipage}
    \end{center}

\end{itemize}

These selection criteria ensure that $D_{\text{redundant}}$ captures responses exhibiting excessive verbosity and hesitation, while $D_{\text{concise}}$ represents efficient and direct responses, providing a clear contrast for computing the steering direction representing overthinking. To ensure high-quality data, we retain only 500 samples for each dataset after applying the selection criteria and double checking. For computing the steering direction, we follow \cite{zou2023representation} to sample 100 samples from each dataset and employ the IsolationForest algorithm to filter out outliers. For manifold subspace estimation, we utilize the entire set of 500 samples from each dataset to capture the full representational structure.

\subsection{Baseline Methods}
As stated in \cref{sec:setting}, we select two latest baselines, Dynasor~\cite{fu2024efficiently} and SEAL~\cite{chen2025seal}, for their ability to preserve the original accuracy in reasoning tasks.  Below, we detail the specific settings for them:

\textbf{General Setting.} All large reasoning models adopt the official recommended settings with a temperature of 0.6, top-p of 0.95, and a maximum length of 16k tokens.

\textbf{Dynasor.} We adopt the official settings for Dynasor. The configuration probes the model every 32 tokens with a ``Probe-In-The-Middle'' technique and injects a ``Final Answer'' prompt at each iteration to ensure complete solutions upon early termination. Generation stops when the Certaindex metric ($\tilde{H}$) exceeds a predefined confidence threshold. To be aware, Dynasor's early stopping often omits the problem-solving process in the final answer, which is impractical for real-world applications. Thus, we require the model to provide a complete solution in the final answer upon stopping.

\textbf{SEAL.} We adopt the official settings for SEAL~\cite{chen2025seal}, using 1k training samples from the Math dataset~\cite{lightman2023lets} to extract the reasoning steering vector. Reasoning processes are segmented into thoughts using ``\textbackslash n\textbackslash n'' delimiters, classified as execution, reflection, or transition via keyword-based rules (e.g., ``Alternatively'' for transition, ``Wait'' for reflection). The steering vector is computed at layer 20 as $ S = \bar{H}_E - \bar{H}_{RT} $, where $\bar{H}_E$ and $\bar{H}_{RT}$ are average representations of execution and reflection/transition thoughts, respectively. During greedy decoding, hidden states of ``\textbackslash n\textbackslash n'' tokens at layer 20 are adjusted as $\tilde{H} = H + 1.0 \cdot S$.

\section{Proofs}
\label{sec:prove}

\subsection{Proof of Theorem \ref{thm:error_amplification}}
\begin{proof}
We derive the expected noise norm of the interference component $\mathbf{r}_{\textit{other}}$, the part of the overthinking direction $\mathbf{r}^{(l^*)}$ in the orthogonal complement $\mathcal{M}^\perp$ of the low-dimensional manifold $\mathcal{M}$. The theorem states:
\begin{equation*}
\mathbb{E}[\|\mathbf{r}_{\textit{other}}\|_2^2] = \text{tr}\left( (\mathbf{I} - \mathbf{P}_{\mathcal{M}}) \mathbf{\Sigma}_{\text{noise}}^{(l)} \right), \quad \mathbf{\Sigma}_{\text{noise}}^{(l)} = \frac{\mathbf{C}^{(l)}}{|D_{\text{redundant}}|} + \frac{\mathbf{C}^{(l)}}{|D_{\text{concise}}|},
\end{equation*}
where $\mathbf{P}_{\mathcal{M}} = \mathbf{U}^{(l)}[:,1:k] (\mathbf{U}^{(l)}[:,1:k])^\top$, and $\mathbf{U}^{(l)}[:,1:k]$ are the top-$k$ principal components of the activation covariance $\mathbf{C}^{(l)}$. We build on prior findings that $\mathcal{M}$ is low-dimensional, identified via PCA on $D_{\text{reasoning}} = D_{\text{redundant}} \cup D_{\text{concise}}$, with $k=10$ capturing over 70\% of the variance, validating the linear manifold assumption.

\textbf{Step 1: Define the overthinking direction $\mathbf{r}^{(l^*)}$.}
Per \cref{eq:overthinking_direction}, $\mathbf{r}^{(l^*)} = \mathbf{r}_{\textit{overthinking}} + \mathbf{r}_{\textit{other}}$, where $\mathbf{r}_{\textit{overthinking}} \in \mathcal{M}$ captures the shift between redundant and concise reasoning, and $\mathbf{r}_{\textit{other}} \in \mathcal{M}^\perp$ is interference. We model:
\begin{equation*}
\mathbf{r}^{(l^*)} = \frac{1}{|D_{\text{redundant}}|} \sum_{x_i \in D_{\text{redundant}}} \mathbf{h}^{(l)}(x_i) - \frac{1}{|D_{\text{concise}}|} \sum_{x_i \in D_{\text{concise}}} \mathbf{h}^{(l)}(x_i).
\end{equation*}
Assume activations $\mathbf{h}^{(l)}(x_i) \sim \mathcal{N}(\boldsymbol{\mu}_{\text{set}}, \mathbf{C}^{(l)})$, with $\boldsymbol{\mu}_{\text{redundant}}$ or $\boldsymbol{\mu}_{\text{concise}}$ for each dataset, and $\mathbf{C}^{(l)}$ estimated over $D_{\text{reasoning}}$. The covariance is:
\begin{equation*}
\mathbb{E}[\mathbf{r}^{(l^*)} \mathbf{r}^{(l^*)\top}] = \frac{\mathbf{C}^{(l)}}{|D_{\text{redundant}}|} + \frac{\mathbf{C}^{(l)}}{|D_{\text{concise}}|}.
\end{equation*}

\textbf{Step 2: Define $\mathcal{M}$ and derive $\mathbf{I} - \mathbf{P}_{\mathcal{M}}$.}
The manifold $\mathcal{M}$ is spanned by the top-$k$ eigenvectors of $\mathbf{C}^{(l)} = \frac{1}{N-1} \mathbf{A}^{(l)} (\mathbf{A}^{(l)} - \bar{\mathbf{A}}^{(l)})^\top$, where $\mathbf{A}^{(l)} = [\mathbf{h}^{(l)}(x_1), \dots, \mathbf{h}^{(l)}(x_N)]$, and $\bar{\mathbf{A}}^{(l)} = \frac{1}{N} \sum_{i=1}^N \mathbf{h}^{(l)}(x_i)$. The eigendecomposition $\mathbf{C}^{(l)} = \mathbf{U}^{(l)} \mathbf{\Lambda}^{(l)} (\mathbf{U}^{(l)})^\top$ yields $\mathbf{U}^{(l)}[:,1:k]$, and:
\begin{equation*}
\mathbf{P}_{\mathcal{M}} = \mathbf{U}^{(l)}[:,1:k] (\mathbf{U}^{(l)}[:,1:k])^\top.
\end{equation*}
The projection onto $\mathcal{M}^\perp$ is $\mathbf{I} - \mathbf{P}_{\mathcal{M}}$, as it removes the $\mathcal{M}$-component. Since $\mathbf{U}^{(l)}[:,1:k]$ is orthonormal, $\mathbf{P}_{\mathcal{M}}$ is idempotent and symmetric, so:
\begin{equation*}
(\mathbf{I} - \mathbf{P}_{\mathcal{M}})^2 = \mathbf{I} - \mathbf{P}_{\mathcal{M}}, \quad (\mathbf{I} - \mathbf{P}_{\mathcal{M}})^\top = \mathbf{I} - \mathbf{P}_{\mathcal{M}}.
\end{equation*}
PCA’s linear basis ensures $\mathcal{M}^\perp$ captures the $d - k$ dimensions of noise, critical when $d \gg k$.

\textbf{Step 3: Define $\mathbf{r}_{\textit{other}}$.}
Since $\mathbf{r}_{\textit{overthinking}} \in \mathcal{M}$, the interference is:
\begin{equation*}
\mathbf{r}_{\textit{other}} = (\mathbf{I} - \mathbf{P}_{\mathcal{M}}) \mathbf{r}^{(l^*)}.
\end{equation*}
This isolates noise in $\mathcal{M}^\perp$, which disrupts normal abilities due to high-dimensional computation.

\textbf{Step 4: Compute the squared norm.}
Calculate:
\begin{equation*}
\|\mathbf{r}_{\textit{other}}\|_2^2 = [(\mathbf{I} - \mathbf{P}_{\mathcal{M}}) \mathbf{r}^{(l^*)}]^\top (\mathbf{I} - \mathbf{P}_{\mathcal{M}}) \mathbf{r}^{(l^*)} = \mathbf{r}^{(l^*)\top} (\mathbf{I} - \mathbf{P}_{\mathcal{M}}) \mathbf{r}^{(l^*)},
\end{equation*}
using the idempotence of $\mathbf{I} - \mathbf{P}_{\mathcal{M}}$.

\textbf{Step 5: Take the expectation.}
Compute:
\begin{equation*}
\mathbb{E}[\|\mathbf{r}_{\textit{other}}\|_2^2] = \mathbb{E}[\mathbf{r}^{(l^*)\top} (\mathbf{I} - \mathbf{P}_{\mathcal{M}}) \mathbf{r}^{(l^*)}] = \text{tr}((\mathbf{I} - \mathbf{P}_{\mathcal{M}}) \mathbb{E}[\mathbf{r}^{(l^*)} \mathbf{r}^{(l^*)\top}]).
\end{equation*}
Substitute:
\begin{equation*}
\mathbb{E}[\mathbf{r}^{(l^*)} \mathbf{r}^{(l^*)\top}] = \mathbf{\Sigma}_{\text{noise}}^{(l)} = \frac{\mathbf{C}^{(l)}}{|D_{\text{redundant}}|} + \frac{\mathbf{C}^{(l)}}{|D_{\text{concise}}|}.
\end{equation*}
Thus:
\begin{equation*}
\mathbb{E}[\|\mathbf{r}_{\textit{other}}\|_2^2] = \text{tr}\left( (\mathbf{I} - \mathbf{P}_{\mathcal{M}}) \mathbf{\Sigma}_{\text{noise}}^{(l)} \right).
\end{equation*}
\end{proof}

\subsection{Proof of Theorem \ref{thm:activation_shift}}
\begin{proof}
We derive the mean activation shift $\Delta \boldsymbol{\mu}^{(l)}$ at layer $l$ due to the intervention (applied as in \cref{eq:intervention_decomp}.) along the overthinking direction $\mathbf{r}^{(l^*)}$, showing its norm is proportional to $\alpha \|\mathbf{r}_{\textit{other}}\|_2$, and establish the layer-wise amplification of the shift at layer $l+1$. The theorem builds on \cref{thm:error_amplification}.
\textbf{Step 1: Derive the mean activation shift.}
The intervention at layer $l$ is:
\begin{equation*}
\mathbf{h}^{(l)'}(x_i) = \mathbf{h}^{(l)}(x_i) - \alpha [(\mathbf{r}^{(l^*)})^\top \mathbf{h}^{(l)}(x_i)] \mathbf{r}^{(l^*)},
\end{equation*}
with $\alpha > 0$. The mean activation before intervention is:
\begin{equation*}
\boldsymbol{\mu}^{(l)} = \frac{1}{N} \sum_{i=1}^N \mathbf{h}^{(l)}(x_i),
\end{equation*}
and post-intervention:
\begin{equation*}
\boldsymbol{\mu}^{(l)'} = \frac{1}{N} \sum_{i=1}^N \mathbf{h}^{(l)'}(x_i) = \frac{1}{N} \sum_{i=1}^N \left( \mathbf{h}^{(l)}(x_i) - \alpha [(\mathbf{r}^{(l^*)})^\top \mathbf{h}^{(l)}(x_i)] \mathbf{r}^{(l^*)} \right).
\end{equation*}
Compute:
\begin{equation*}
\boldsymbol{\mu}^{(l)'} = \boldsymbol{\mu}^{(l)} - \alpha \frac{1}{N} \sum_{i=1}^N [(\mathbf{r}^{(l^*)})^\top \mathbf{h}^{(l)}(x_i)] \mathbf{r}^{(l^*)}.
\end{equation*}
The mean shift is:
\begin{equation*}
\Delta \boldsymbol{\mu}^{(l)} = \boldsymbol{\mu}^{(l)'} - \boldsymbol{\mu}^{(l)} = -\alpha \frac{1}{N} \sum_{i=1}^N [(\mathbf{r}^{(l^*)})^\top \mathbf{h}^{(l)}(x_i)] \mathbf{r}^{(l^*)},
\end{equation*}
matching the first part of \cref{eq:activation_shift}.

\textbf{Step 2: Decompose the shift and isolate $\mathbf{r}_{\textit{other}}$ contribution.}
Since $\mathbf{r}^{(l^*)} = \mathbf{r}_{\mathcal{M}} + \mathbf{r}_{\textit{other}}$ with $\mathbf{r}_{\mathcal{M}} \perp \mathbf{r}_{\textit{other}}$, we can decompose:
\begin{equation*}
(\mathbf{r}^{(l^*)})^\top \mathbf{h}^{(l)}(x_i) = (\mathbf{r}_{\mathcal{M}})^\top \mathbf{h}^{(l)}(x_i) + (\mathbf{r}_{\textit{other}})^\top \mathbf{h}^{(l)}(x_i).
\end{equation*}
Thus:
\begin{align*}
\Delta \boldsymbol{\mu}^{(l)} &= -\alpha \frac{1}{N} \sum_{i=1}^N [(\mathbf{r}_{\mathcal{M}})^\top \mathbf{h}^{(l)}(x_i) + (\mathbf{r}_{\textit{other}})^\top \mathbf{h}^{(l)}(x_i)] (\mathbf{r}_{\mathcal{M}} + \mathbf{r}_{\textit{other}}) \\
&= -\alpha \frac{1}{N} \sum_{i=1}^N [(\mathbf{r}_{\mathcal{M}})^\top \mathbf{h}^{(l)}(x_i)] \mathbf{r}_{\mathcal{M}} - \alpha \frac{1}{N} \sum_{i=1}^N [(\mathbf{r}_{\mathcal{M}})^\top \mathbf{h}^{(l)}(x_i)] \mathbf{r}_{\textit{other}} \\
&\quad - \alpha \frac{1}{N} \sum_{i=1}^N [(\mathbf{r}_{\textit{other}})^\top \mathbf{h}^{(l)}(x_i)] \mathbf{r}_{\mathcal{M}} - \alpha \frac{1}{N} \sum_{i=1}^N [(\mathbf{r}_{\textit{other}})^\top \mathbf{h}^{(l)}(x_i)] \mathbf{r}_{\textit{other}}.
\end{align*}
Let:
\begin{equation*}
s_{\mathcal{M}} = \frac{1}{N} \sum_{i=1}^N [(\mathbf{r}_{\mathcal{M}})^\top \mathbf{h}^{(l)}(x_i)], \quad s_{\textit{other}} = \frac{1}{N} \sum_{i=1}^N [(\mathbf{r}_{\textit{other}})^\top \mathbf{h}^{(l)}(x_i)].
\end{equation*}
Then:
\begin{equation*}
\Delta \boldsymbol{\mu}^{(l)} = -\alpha s_{\mathcal{M}} \mathbf{r}_{\mathcal{M}} - \alpha s_{\mathcal{M}} \mathbf{r}_{\textit{other}} - \alpha s_{\textit{other}} \mathbf{r}_{\mathcal{M}} - \alpha s_{\textit{other}} \mathbf{r}_{\textit{other}}.
\end{equation*}
Since $\mathbf{r}_{\mathcal{M}} \perp \mathbf{r}_{\textit{other}}$:
\begin{align*}
\|\Delta \boldsymbol{\mu}^{(l)}\|_2^2 &= \alpha^2 \|s_{\mathcal{M}} \mathbf{r}_{\mathcal{M}} + s_{\textit{other}} \mathbf{r}_{\mathcal{M}}\|_2^2 + \alpha^2 \|s_{\mathcal{M}} \mathbf{r}_{\textit{other}} + s_{\textit{other}} \mathbf{r}_{\textit{other}}\|_2^2 \\
&= \alpha^2 (s_{\mathcal{M}} + s_{\textit{other}})^2 \|\mathbf{r}_{\mathcal{M}}\|_2^2 + \alpha^2 (s_{\mathcal{M}} + s_{\textit{other}})^2 \|\mathbf{r}_{\textit{other}}\|_2^2.
\end{align*}
Let $s = s_{\mathcal{M}} + s_{\textit{other}}$. Then:
\begin{equation*}
\|\Delta \boldsymbol{\mu}^{(l)}\|_2 = \alpha |s| \sqrt{\|\mathbf{r}_{\mathcal{M}}\|_2^2 + \|\mathbf{r}_{\textit{other}}\|_2^2}.
\end{equation*}
By \cref{thm:error_amplification}, when the error component $\mathbf{r}_{\textit{other}}$ is present (i.e., $\|\mathbf{r}_{\textit{other}}\|_2 > 0$), it contributes to the total norm. The dominant term depends on the relative magnitudes of $\|\mathbf{r}_{\mathcal{M}}\|_2$ and $\|\mathbf{r}_{\textit{other}}\|_2$. Assuming $\mathbf{h}^{(l)}(x_i) \sim \mathcal{N}(\boldsymbol{\mu}_{\text{set}}, \mathbf{C}^{(l)})$ and $|s|$ is a positive constant, we obtain:
\begin{equation*}
\|\Delta \boldsymbol{\mu}^{(l)}\|_2 \propto \alpha \|\mathbf{r}_{\textit{other}}\|_2,
\end{equation*}
when $\|\mathbf{r}_{\textit{other}}\|_2$ dominates, completing \cref{eq:activation_shift}.

\textbf{Step 3: Derive the layer-wise amplification from $\mathbf{r}_{\textit{other}}$.}
For layer $l+1$, the activation is:
\begin{equation*}
\mathbf{h}^{(l+1)}(x_i) = \sigma\left( \mathbf{W}^{(l+1)} \text{Attn}(\mathbf{h}^{(l)}(x_i)) \right),
\end{equation*}
and post-intervention:
\begin{equation*}
\mathbf{h}^{(l+1)'}(x_i) = \sigma\left( \mathbf{W}^{(l+1)} \text{Attn}(\mathbf{h}^{(l)'}(x_i)) \right),
\end{equation*}
where $\mathbf{W}^{(l+1)}$ combines MLP and attention weights, $\text{Attn}$ is the attention mechanism, and $\sigma$ is GeLU. The mean shift is:
\begin{equation*}
\Delta \boldsymbol{\mu}^{(l+1)} = \boldsymbol{\mu}^{(l+1)'} - \boldsymbol{\mu}^{(l+1)}, \quad \boldsymbol{\mu}^{(l+1)'} = \frac{1}{N} \sum_{i=1}^N \mathbf{h}^{(l+1)'}(x_i), \quad \boldsymbol{\mu}^{(l+1)} = \frac{1}{N} \sum_{i=1}^N \mathbf{h}^{(l+1)}(x_i).
\end{equation*}
To isolate the $\mathbf{r}_{\textit{other}}$ contribution, decompose the single-input shift at layer $l$:
\begin{align*}
\Delta \mathbf{h}^{(l)}(x_i) &= \mathbf{h}^{(l)'}(x_i) - \mathbf{h}^{(l)}(x_i) \\
&= -\alpha [(\mathbf{r}^{(l^*)})^\top \mathbf{h}^{(l)}(x_i)] \mathbf{r}^{(l^*)} \\
&= -\alpha [(\mathbf{r}_{\mathcal{M}})^\top \mathbf{h}^{(l)}(x_i) + (\mathbf{r}_{\textit{other}})^\top \mathbf{h}^{(l)}(x_i)] (\mathbf{r}_{\mathcal{M}} + \mathbf{r}_{\textit{other}}).
\end{align*}
The norm is:
\begin{align*}
\|\Delta \mathbf{h}^{(l)}(x_i)\|_2 &= \alpha |(\mathbf{r}^{(l^*)})^\top \mathbf{h}^{(l)}(x_i)| \|\mathbf{r}^{(l^*)}\|_2 \\
&= \alpha |(\mathbf{r}_{\mathcal{M}})^\top \mathbf{h}^{(l)}(x_i) + (\mathbf{r}_{\textit{other}})^\top \mathbf{h}^{(l)}(x_i)| \sqrt{\|\mathbf{r}_{\mathcal{M}}\|_2^2 + \|\mathbf{r}_{\textit{other}}\|_2^2}.
\end{align*}
The component from $\mathbf{r}_{\textit{other}}$ can be isolated by considering its contribution:
\begin{equation*}
\|\Delta \mathbf{h}^{(l)}(x_i)\|_2 \geq \alpha |(\mathbf{r}_{\textit{other}})^\top \mathbf{h}^{(l)}(x_i)| \|\mathbf{r}_{\textit{other}}\|_2,
\end{equation*}
when the $\mathbf{r}_{\textit{other}}$ term is significant. Propagate to layer $l+1$:
\begin{equation*}
\Delta \mathbf{h}^{(l+1)}(x_i) = \mathbf{h}^{(l+1)'}(x_i) - \mathbf{h}^{(l+1)}(x_i) \approx \sigma'\left( \mathbf{W}^{(l+1)} \text{Attn}'(\mathbf{h}^{(l)}(x_i)) \Delta \mathbf{h}^{(l)}(x_i) \right),
\end{equation*}
where $\text{Attn}'$ and $\sigma'$ are the Jacobians of attention and GeLU. The attention softmax and GeLU have minimum amplification factors $\gamma_{\text{attn}}, \gamma_{\sigma} > 0$, and the linear transformation by $\mathbf{W}^{(l+1)}$ satisfies:
\begin{equation*}
\|\mathbf{W}^{(l+1)} \mathbf{x}\|_2 \geq \sigma_{\text{min}}(\mathbf{W}^{(l+1)}) \|\mathbf{x}\|_2.
\end{equation*}
Thus:
\begin{equation*}
\|\Delta \mathbf{h}^{(l+1)}(x_i)\|_2 \geq \gamma_{\text{attn}} \gamma_{\sigma} \sigma_{\text{min}}(\mathbf{W}^{(l+1)}) \|\Delta \mathbf{h}^{(l)}(x_i)\|_2.
\end{equation*}
Focusing on the $\mathbf{r}_{\textit{other}}$ contribution:
\begin{equation*}
\|\Delta \mathbf{h}^{(l+1)}(x_i)\|_2 \geq \gamma_{\text{attn}} \gamma_{\sigma} \sigma_{\text{min}}(\mathbf{W}^{(l+1)}) \alpha |(\mathbf{r}_{\textit{other}})^\top \mathbf{h}^{(l)}(x_i)| \|\mathbf{r}_{\textit{other}}\|_2.
\end{equation*}
The mean shift norm is:
\begin{equation*}
\|\Delta \boldsymbol{\mu}^{(l+1)}\|_2 = \left\| \frac{1}{N} \sum_{i=1}^N \Delta \mathbf{h}^{(l+1)}(x_i) \right\|_2.
\end{equation*}
Assume the layer-wise propagation amplifies the previous shift by $\gamma > 1$, reflecting attention and non-linear effects across layers. Combining the amplification of the existing shift and the new $\mathbf{r}_{\textit{other}}$ contribution:
\begin{equation*}
\|\Delta \boldsymbol{\mu}^{(l+1)}\|_2 \geq \gamma \|\Delta \boldsymbol{\mu}^{(l)}\|_2 + \alpha \gamma_{\text{attn}} \gamma_{\sigma} \sigma_{\text{min}}(\mathbf{W}^{(l+1)}) |(\mathbf{r}_{\textit{other}})^\top \mathbf{h}^{(l)}(x_i)| \|\mathbf{r}_{\textit{other}}\|_2,
\end{equation*}
matching \cref{eq:layer_amplification}. This shows that the $\mathbf{r}_{\textit{other}}$ component causes layer-wise amplification through both the accumulated shift (first term) and the direct contribution at each layer (second term).

\textbf{Step 4: Analyze the amplification mechanism.}
The amplification factors $\gamma > 1$, $\gamma_{\text{attn}}, \gamma_{\sigma} > 0$, and non-zero $\sigma_{\text{min}}(\mathbf{W}^{(l+1)})$ ensure that perturbations from $\mathbf{r}_{\textit{other}}$ grow across layers. The first term $\gamma \|\Delta \boldsymbol{\mu}^{(l)}\|_2$ represents the propagation of accumulated shift, while the second term represents the fresh perturbation introduced at layer $l+1$ due to $\mathbf{r}_{\textit{other}}$. This dual mechanism ensures the shift grows across layers, disrupting the model's normal abilities.
\end{proof}

\section{Hyperparameter Tuning}
\label{sec:hyper_more}
\begin{figure}[!h]
    \centering
    \includegraphics[width=\linewidth]{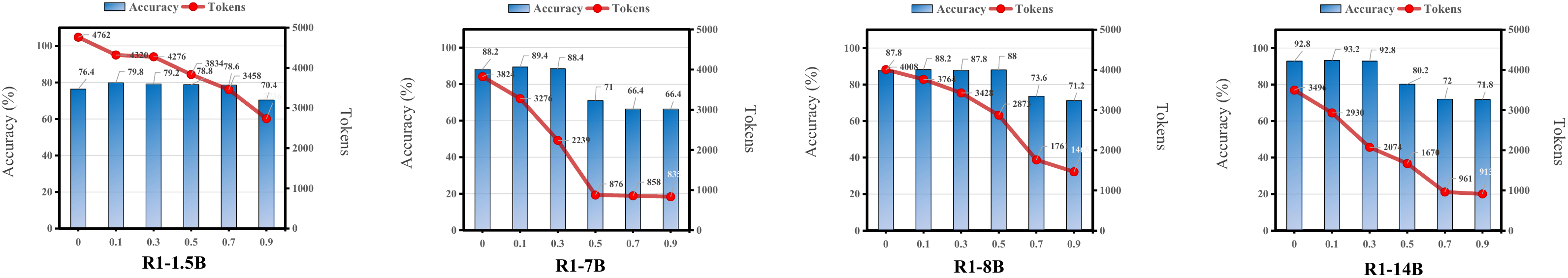}
    \caption{Impact of intervention strength $\alpha$ on the token reduction and accuracy of R1-1.5B, R1-7B, R1-8B, and R1-14B on the MATH500 dataset}
    \label{fig:hyper_more}
\end{figure}

In this section, we present the results of tuning the intervention strength $\alpha$ across four models: DeepSeek-R1-Distill-Qwen-1.5B, DeepSeek-R1-Distill-Qwen-7B, DeepSeek-R1-Distill-Llama-8B, and DeepSeek-R1-Distill-Qwen-14B on MATH500~\cite{lightman2023lets}. As shown in \cref{fig:hyper_more}, to achieve an optimal balance between efficiency and accuracy, we ultimately select $\alpha = 0.7$ for R1-1.5B, $\alpha = 0.3$ for R1-7B, $\alpha = 0.5$ for R1-8B, and $\alpha = 0.3$ for R1-14B.

\section{Layer Selection for Manifold Steering}
\label{sec:layer_selection}

The selection of intervention layers is critical for the effectiveness of Manifold Steering. We conduct a layer-wise analysis across multiple model sizes to determine the optimal intervention points. As shown in the tables below, we evaluate the performance across different layers by measuring accuracy and tokens on the MATH500. The results demonstrate that later layers consistently achieve better performance: Layer 27 for R1-1.5B and R1-7B, Layer 31  for R1-8B, and Layer 47  for R1-14B.

\begin{table}[!h]
\centering
\caption{Layer-wise performance analysis for R1-1.5B on MATH500.}
\label{tab:layer_selection_1.5b}
\resizebox{\linewidth}{!}{
\begin{tabular}{lcccccccc}
\toprule
 & \textbf{Vanilla} & \textbf{Layer 1} & \textbf{Layer 5} & \textbf{Layer 10} & \textbf{Layer 15} & \textbf{Layer 20} & \textbf{Layer 25} & \textbf{Layer 27} \\
\midrule
Accuracy (\%) & 76.4 & 76.6 & 77.0 & 76.4 & 57.6 & 74.8 & 67.4 & 78.6 \\
\# Tokens & 4762 & 4472 & 4434 & 4223 & 1469 & 3930 & 1179 & 3458 \\
\bottomrule
\end{tabular}}
\end{table}

\begin{table}[!h]
\centering
\caption{Layer-wise performance analysis for R1-7B on MATH500.}
\label{tab:layer_selection_7b}
\resizebox{\linewidth}{!}{
\begin{tabular}{lcccccccc}
\toprule
 & \textbf{Vanilla} & \textbf{Layer 1} & \textbf{Layer 5} & \textbf{Layer 10} & \textbf{Layer 15} & \textbf{Layer 20} & \textbf{Layer 25} & \textbf{Layer 27} \\
\midrule
Accuracy (\%) & 88.2 & 88.4 & 88.0 & 88.2 & 84.4 & 80.6 & 72.2 & 88.4 \\
\# Tokens & 3824 & 3685 & 3665 & 3701 & 2713 & 1906 & 1070 & 2239 \\
\bottomrule
\end{tabular}}
\end{table}

\begin{table}[!h]
\centering
\caption{Layer-wise performance analysis for R1-8B on MATH500.}
\label{tab:layer_selection_8b}
\resizebox{\linewidth}{!}{
\begin{tabular}{lccccccccc}
\toprule
 & \textbf{Vanilla} & \textbf{Layer 1} & \textbf{Layer 5} & \textbf{Layer 10} & \textbf{Layer 15} & \textbf{Layer 20} & \textbf{Layer 25} & \textbf{Layer 30} & \textbf{Layer 31} \\
\midrule
Accuracy (\%) & 87.8 & 87.2 & 87.8 & 88.2 & 75.6 & 86.4 & 71.8 & 87.6 & 88.0 \\
\# Tokens & 4009 & 3896 & 3820 & 3654 & 2950 & 3280 & 1856 & 2975 & 2873 \\
\bottomrule
\end{tabular}}
\end{table}

\begin{table}[!h]
\centering
\caption{Layer-wise performance analysis for R1-14B on MATH500.}
\label{tab:layer_selection_14b}
\resizebox{\linewidth}{!}{
\begin{tabular}{lcccccccccccc}
\toprule
 & \textbf{Vanilla} & \textbf{Layer 1} & \textbf{Layer 5} & \textbf{Layer 10} & \textbf{Layer 15} & \textbf{Layer 20} & \textbf{Layer 25} & \textbf{Layer 30} & \textbf{Layer 35} & \textbf{Layer 40} & \textbf{Layer 45} & \textbf{Layer 47} \\
\midrule
Accuracy (\%) & 92.8 & 92.4 & 92.4 & 92.0 & 92.2 & 92.6 & 89.8 & 80.4 & 87.4 & 84.6 & 82.4 & 92.8 \\
\# Tokens & 3496 & 3384 & 3420 & 3095 & 2958 & 2857 & 2398 & 1814 & 2207 & 1836 & 1625 & 2074 \\
\bottomrule
\end{tabular}}
\end{table}

\section{Time Latency Analysis}
\label{sec:compute}
In this section, we analyze the time latency for the DeepSeek-R1-Distill-Qwen-7B model on the Math500 dataset~\cite{lightman2023lets}, comparing our approach with Dynasor~\cite{fu2024efficiently} and SEAL~\cite{chen2025seal}. All experiments are conducted on an Ubuntu 22.04 system with A800 GPUs. We find that Dynasor exhibits the significantly longest time latency, which is reasonable due to its frequent probing of intermediate states and its unsuitability for parallel processing of large reasoning models.
For SEAL, although both SEAL and our method introduce negligible additional computational cost, SEAL’s token reduction is less effective than ours, resulting in higher time latency.

\begin{table}[!h]
\centering
\caption{Average Time Latency on Math500 for different overthinking-mitigation methods in R1-7B.}
\begin{tabular}{ccccc}
\toprule[1.5pt]
Methods   & Original & Dynasor & SEAL & Ours \\
\midrule
Time Latency (s) &   1.74       &   39.89      &  1.37    & 1.05    \\
\bottomrule[1.5pt]
\end{tabular}
\label{tab:time_cost}
\end{table}

%%%%%%%%%%%%%%%%%%%%%%%%%%%%%%%%%%%%%%%%%%%%%%%%%%%%%%%%%%%%

\end{document}